\documentclass[11pt,letterpaper]{article}
\usepackage{emnlp2017}
\usepackage{times}
\usepackage{latexsym}
\usepackage{graphicx}
\usepackage{hyperref}
\usepackage{paralist}
\usepackage{tikz-dependency}
\usepackage{booktabs} 
\usepackage{tablefootnote}
\usepackage{todonotes}

\emnlpfinalcopy

\title{To Normalize, or Not to Normalize:\\The Impact of Normalization on Part-of-Speech Tagging}
\author
{
	\begin{tabular}{ccc}
	Rob van der Goot & Barbara Plank & Malvina Nissim \\
	\end{tabular}
	\\
    Center for Language and Cognition, University of Groningen, The Netherlands\\
	{\tt \{r.van.der.goot,b.plank,m.nissim\}@rug.nl}
}

\date{}

\begin{document}
\maketitle
\begin{abstract}
 Does normalization help Part-of-Speech (POS) tagging accuracy on noisy, non-canonical data?
To the best of our knowledge, little is known on the actual impact of normalization in a real-world scenario, where gold error detection is not available.  We investigate the effect of automatic normalization on POS tagging of tweets.
We also compare normalization to strategies that leverage large amounts of unlabeled data kept in its raw form.  Our results show that normalization helps, but does not add  consistently beyond just word embedding layer initialization. The latter approach yields a tagging model that is competitive with a Twitter state-of-the-art tagger.
\end{abstract} 

\section{Introduction}
Non-canonical data poses a series of challenges to Natural Language Processing, 
as reflected in large performance drops documented in a variety of tasks, e.g., on POS tagging~\cite{gimpel:ea:2011,hovy:ea:2014}, parsing~\cite{mcclosky:2010,foster:ea:2011} and named entity recognition~\cite{Ritter11}. In this paper we focus on POS tagging and on a particular source of non-canonical language, namely Twitter data.

One obvious way to tackle the problem of processing non-canonical data is to build taggers that are specifically tailored to such text. A prime example is the ARK POS tagger, designed to process English Twitter data \cite{gimpel:ea:2011,owoputi-EtAl:2013:NAACL-HLT}, on which it achieves state-of-the-art results. One drawback of such an approach is that non-canonical data is not all of the same kind, so that for non-canonical non-Twitter data or even collections of Twitter samples from different times, a new specifically dedicated tool needs to be created.

The alternative route is to take a general purpose state-of-the-art POS tagger and adapt it to successfully tag non-canonical data. 
In the case of Twitter, one way to go about this is \textit{lexical normalization}. It is the task of detecting ``ill-formed'' words~\cite{han-baldwin:2011:ACL-HLT2011} and replacing them with their canonical counterpart.
To illustrate why this might help, consider the following tweet: ``new pix comming tomoroe''.  An off-the-shelf system such as the Stanford NLP suite\footnote{\url{http://nlp.stanford.edu:8080/parser/index.jsp}, accessed June 1, 2017.} makes several mistakes on the raw input, e.g.,  the verb `comming' as well as the plural noun `pix' are tagged as singular noun. Instead,  its normalized form is analyzed correctly,  as shown in Figure~\ref{fig:example}.

\begin{figure}
\centering
	\begin{tabular}{l l l l}
	JJ 	& NN & NN & NNS\\
	new & pix & comming & tomoroe\\
	JJ 	& NNS & VBG & NN\\
	new & pictures & coming & tomorrow\\
	\end{tabular}
    \label{fig:example}
	\caption{Example tweet from the test data, raw and normalized form, tagged with Stanford NLP.}
\end{figure} 

While being a promising direction, we see at least two issues with the assessment of normalization as a successful step in POS tagging non-canonical text. Firstly, normalization experiments are usually carried out assuming that the tokens to be normalized are already detected (\textit{gold error detection}). Thus little is known on how normalization impacts tagging accuracy in a real-world scenario (\textit{not} assuming gold error detection). Secondly, normalization is one way to go about processing non-canonical data, but not the only one~\cite{eisenstein:2013:NAACL-HLT,Plank:2016:KONVENS}. Indeed, alternative approaches include leveraging the abundance of \textit{unlabeled data} kept in its raw form. For instance, such data can be exploited with semi-supervised learning methods \cite{abney2007}. The advantage of this approach is that portability could be successful also towards domains where normalization is not necessary or crucial. These observations lead us to the following research questions: 

\begin{enumerate}
\item[Q1] In a real-world setting, without assuming gold error detection, does normalization help in POS tagging of tweets?
\item[Q2] In the context of POS tagging, is it more beneficial to normalize input data or is it better to work with raw data and exploit large amounts of it in a semi-supervised setting?
\item[Q3] To what extent are normalization and semi-supervised approaches complementary? 
\end{enumerate}
To answer these questions, we run a battery of experiments that evaluate different approaches. Specifically:

\begin{enumerate} 
\item We study the impact of normalization on POS tagging in a realistic setup, i.e., we compare normalizing only unknown words, or words for which we know they need correction; we compare this with a fully automatic normalization model (Section~\ref{sec:normalization}).

\item We evaluate the impact of leveraging large amounts of unlabeled data using two approaches: a) deriving various word representations, and studying their effect for model initialization (Section~\ref{sec:embeddings}); b) applying a bootstrapping approach based on self-training to automatically derive labeled training data, evaluating a range of a-priori data selection mechanisms
(Section~\ref{sec:selftraining}).

\item We experiment with combining the most promising methods from both directions,  to gain insights on their potential complementarity (Section~\ref{sec:analysis}).
\end{enumerate}

\section{Experimental Setup}
\label{sec:setup}

We run two main sets of POS tagging experiments. In the first one, we use normalization in a variety of settings (see Section~\ref{sec:normalization}). In the second one, we leverage large amounts of unlabeled data that does not undergo any normalization but is used as training in a semi-supervised setting (Section~\ref{sec:SSL}). For all experiments we use existing datasets as well as newly created resources, cf.\ Section~\ref{sec:data}. The POS model used is described in Section~\ref{sec:bilty}.

\subsection{Data}
\label{sec:data}
\begin{figure}
\centering
\includegraphics[width=\columnwidth]{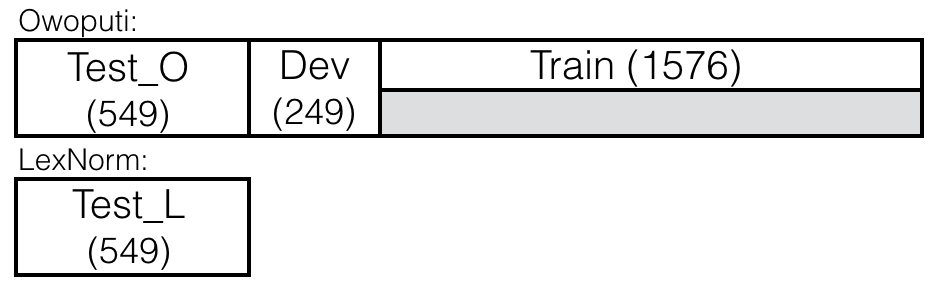}
\caption{Labeled data for POS and normalization. Gray area: no gold normalization layer available.}
\label{fig:data}
\end{figure}

The annotated datasets used in this study originate from two sources: ~\citet{owoputi-EtAl:2013:NAACL-HLT} and~\citet{han-baldwin:2011:ACL-HLT2011}, which we will refer to as \textsc{Owoputi} and \textsc{LexNorm}, respectively.
All datasets used in this study are annotated with the 26 Twitter tags as described in~\cite{gimpel:ea:2011}.\footnote{Some tags are rare, like M and Y. In fact, M occurs only once in \textsc{Test\_L}; Y never occurs in \textsc{Dev} and only once in \textsc{Test\_L} and three times in \textsc{Test\_O}. Therefore our confusion matrices (over \textsc{Dev} and \textsc{Test\_O}, respectively) have different number of labels on the axes.}
\textsc{Owoputi} was originally annotated with POS labels, 
whereas \textsc{LexNorm} was solely annotated for normalization. \newcite{li2015joint} added a POS tag layer to the \textsc{LexNorm} corpus, and a normalization layer to 798 Tweets from \textsc{Owoputi}, which we split into a separate \textsc{Dev} and \textsc{Test} part of 249 and 549 Tweets, respectively, keeping the original POS labels.
We use \textsc{Dev} throughout all experiments during development, and test our final best system on the held-out test sets (both containing 549 tweets). An illustration of the data is given in Figure~\ref{fig:data}.

For the different improvements to our baseline tagger, we  need raw data from the target domain (Twitter). In addition, the normalization model needs unlabeled canonical data. We use a snapshot of  English Wikipedia as unlabeled canonical data source.
To get raw data for the social media domain, we collected Tweets during the whole year of 2016 by means of the Twitter API. 
We only collected Tweets containing one of the 100 frequent words in the Oxford English Corpus\footnote{\url{https://en.wikipedia.org/wiki/Most_common_words_in_English}} as a rough language filter. This resulted in a dataset of 760,744,676 English Tweets. 
We do some very basic pre-processing in which we replace urls and usernames by $<$URL$>$ and $<$USERNAME$>$, and remove duplicate tweets. Because of different casing strategies, we always apply a simple postprocessing step to `rt' (retweet) tokens.

\subsection{Model} 
\label{sec:bilty}

We use \textsc{Bilty}, an off-the-shelf bi-directional Long Short-Term Memory (bi-LSTM) tagger which utilizes both word and character embeddings~\cite{plank:ea:2016}. 
The tagger is trained on 1,576 training tweets (Section~\ref{sec:data}). We tune the parameters of the POS tagger on the development set to derive the following hyperparameter setup, which we use throughout the rest of the experiments: 10 epochs, 1 bi-LSTM layer, 100 input dimensions for words, 256 for characters, $\sigma$=0.2, \texttt{constant} embeddings initializer,  \texttt{Adam} trainer, and updating embeddings during backpropagation.\footnote{\texttt{Adam} was consistently better than \texttt{sgd} on this small training dataset. More LSTM layers lowered performance.} 

\section{To Normalize}
\label{sec:normalization}
First we evaluate the impact of normalization on the POS tagger.
\subsection{Model}
\label{sec:normSystem}
We use an in-house developed normalization model~\cite{monoise}.\footnote{Available at: \url{https://bitbucket.org/robvanderg/monoise}}
The model is based on the assumption that different normalization problems require different handling. First, since unintentional disfluencies can often be corrected by the use of a spell checker, the normalization model exploits Aspell\footnote{\url{www.aspell.net}}. 
Second, since intentional disfluencies typically have a much larger edit distance, the normalization system uses word embeddings~\cite{mikolov2013efficient};\footnote{Using the tweets from Section~\ref{sec:data} and the following parameters: \texttt{-size 400 -window 1 -negative 5 -sample 1e-4 -iter 5}}
 words close to the non-canonical word in the vector space are considered potential normalization candidates. On top of that, the model uses a lookup list generated from the training data, which works especially well for slang.
 
Features originating from the ranking are combined with uni- and bi-gram probabilities from Wikipedia data as well as from raw Tweets (Section~\ref{sec:data}).
A random forest classifier~\cite{breiman2001random} is then used to rank the candidates for each word.
Note that the original word is also a candidate; this enables the model to handle error detection,
which is not always the case in models of previous work.

We train the normalization model on 2,577 tweets from~\newcite{li-liu:2014:P14-3}. 
Our model~\cite{monoise} achieves state-of-art performance
on the erroneous tokens (using gold error detection) on the LexNorm dataset~\cite{han-baldwin:2011:ACL-HLT2011} as well as state-of-art on another corpus which is usually benchmarked without assuming gold error detection~\cite{baldwin-EtAl:2015:WNUT}. We refer the reader to the paper~\cite{monoise} for further details.

To obtain a more detailed view of the effect of normalization on POS tagging, we investigate four experimental setups:
\begin{itemize}
	\item normalizing only unknown words;
	\item considering all words: the model decides whether a word should be normalized or not;
	\item assuming gold error detection: the model knows which words should be normalized;
	\item gold normalization; we consider this a theoretical upper bound. 
\end{itemize}

Traditionally, normalization is used to make the test data more similar to the train data. Since we train our tagger on the social media domain as well, the normalization of only the test data might actually result in more distance between the train and test data. 
Therefore, we also train the tagger on normalized training data, and on the union of the normalized and the original training data.

\subsection{Results}

\begin{table*}[thb]
\centering
		\begin{tabular}{l | r | r r | r r} \toprule
			$\downarrow$ \textbf{Train}  $\rightarrow$ \textbf{Test}   & \textsc{Raw}	& \textsc{Unk}	& \textsc{All}  & \textsc{GoldED}	& \textsc{Gold}\\
   	 	    \midrule
    	    \textsc{Raw}		& 82.16 ($\pm$.33) & 83.44 ($\pm$.25) & \textbf{84.06} ($\pm$.32) & 84.67 ($\pm$.23) & 86.71 ($\pm$.25)\\
			\midrule
			\textsc{All}		& 80.42 ($\pm$.71) & 81.99 ($\pm$.64) & 83.87 ($\pm$.28) & 84.05 ($\pm$.31) & 86.11 ($\pm$.14) \\
        	\textsc{Union}		& 81.54 ($\pm$.27)  & 83.11 ($\pm$.31)  & 84.04  ($\pm$.34) & 84.42 ($\pm$.24) & 86.35 ($\pm$.17) \\
            \bottomrule
    	\end{tabular}
    \caption{Results of normalization (N) on \textsc{Dev} (macro average and stdev over 5 runs).  \textsc{Raw}: no normalization, \textsc{All}: automatic normalization, \textsc{Unk}: normalize only unknown words, \textsc{GoldED}: use gold error detection, \textsc{Gold}: use gold normalization (Oracle). Row: whether training data is normalized. \textsc{Union} stands for the training set formed by the union of both normalized and original raw data.} 
    \label{tab:norm}
\end{table*}

The effects of the different normalization strategies on the \textsc{Dev} data are shown in Table~\ref{tab:norm}. Throughout the paper we report average accuracies  over 5 runs including standard deviation.

The first row shows the effect of normalization at test-time only.
From these results we can conclude that normalizing all words is beneficial over normalizing only unknown words; this shows that normalization has a positive effect that goes beyond changing unknown words.

The results of using the gold normalization suggest that there is still more to gain by improving the  normalization model. In contrast, the results for gold error detection (\textsc{GoldED}) show that error detection is not the main reason for this difference, since the performance difference between \textsc{All} and \textsc{GoldED} is relatively small compared to the gap with \textsc{Gold}.

Considering the normalization of the training data, we see that it has a negative effect. The table suggests that training on the raw  (non-normalized) training data works best.  Adding normalized data to raw data (\textsc{Union}) does not yield any clear improvement over \textsc{Raw} only, but requires more training time.  For the test data, normalization is instead always beneficial.

To sum up, normalization improved the base tagger by 1.9\% absolute performance on the development data, reaching 84.06\% accuracy.
Overall, our state-of-art normalization model only reaches approximately 50\% of the theoretical upper bound of using gold normalization.
We next investigate whether using large amounts of unlabeled data can help us to obtain a similar effect.

\section{Or Not to Normalize}
\label{sec:SSL}
An alternative option to normalization is to leave the text as is, and exploit very large amounts of raw data via semi-supervised learning. The rationale behind this is the following: provided the size of the data is sufficient, a model can be trained to naturally learn the POS tags of noisy data.
\subsection{Effect of Word Embeddings}
\label{sec:embeddings}
An easy and effective use of word embeddings in neural network approaches is to use them to initialize the word lookup parameters. 

We train a skip-gram word embeddings model using word2vec~\cite{mikolov2013efficient} on 760M tweets (as described in Section~\ref{sec:normSystem}). We also experiment with structured skip-grams
\cite{ling-EtAl:2015:NAACL-HLT}, an adaptation of word2vec which takes word order into account.
It has been shown to be beneficial for syntactically oriented tasks, like POS tagging. Therefore we want to evaluate structured skip-grams as well.

The normalization model uses word embeddings with a window size of 1; we compare this with the default window size of 5 for structured skip-grams.%\footnote{Other parameters are constant: \texttt{-size 100 -negative 5 -sample 1e-4 -iter 5}} 

\paragraph{Results}

\begin{table}[b]
\resizebox{\columnwidth}{!}{
	\begin{tabular}{l ccc}\toprule
    & \multicolumn{2}{c}{\textsc{window size}} \\  
    							&  \textsc{1}	&  \textsc{5} \\ \midrule
    \textsc{skipg.}                   & 88.14 ($\pm$.30) & 87.56 ($\pm$.08)\\
    \textsc{struct.skipg. }        & \textbf{88.51} ($\pm$.24) & 88.11 ($\pm$.49) \\ \bottomrule
    \end{tabular}
    }
    \caption{Accuracy on raw \textsc{Dev}:\ various pre-trained skip-gram embeddings for initialization.}
    \label{tab:embeds}
\end{table}

Table~\ref{tab:embeds} shows the results of using the different skip-gram models for initialization of the word embeddings layer.
Structured skip-grams perform slightly better, confirming earlier findings. 
Using a smaller window is more beneficial, probably because of the fragmented nature of Twitter data. 

Structured skip-grams of window size 1 result in the best embedding model. This results in an improvement from 82.16\% (Table~\ref{tab:norm}) to 88.51\% accuracy. This improvement is considerably larger than what obtained by normalization (82.16). 

\subsection{Effect of Self-training} 
\label{sec:selftraining}
\begin{figure}
	\centering
	\includegraphics[width=.4\textwidth]{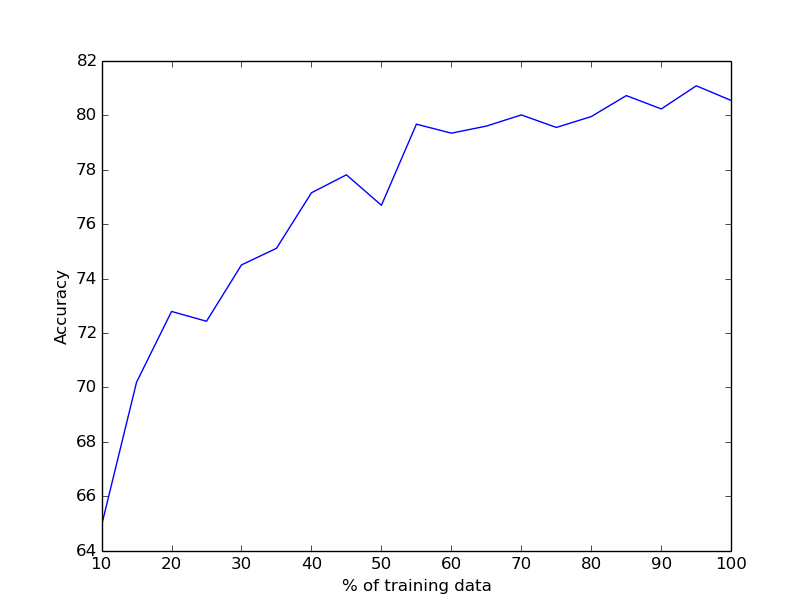}
	\caption{Effect of increasing amounts of training data (100\% training data == 1,576 tweets). \label{fig:learnCurve}}
\end{figure}

We work with a rather small training set, which is all that is available to us in terms of gold data. This is due to the use of an idiosyncratic tagset~\cite{gimpel:ea:2011}. Adding more data could be beneficial to the system. To get an idea of how much effect extra annotated data could potentially have on POS tag accuracy, we plot the performance using smaller amounts of gold training data in Figure~\ref{fig:learnCurve}. We can see that there is still a slight upward trend; however, even when adding manually annotated data, the performance sometimes drop, especially after adding 55\% of the training data.

To create more training data, we use an iterative indelible self-training setup~\cite{abney2007} to obtain automatically labeled data. Specifically: 100 tweets are tagged, they get added to the training data, and after this a new model is trained. 

While we do not adopt any filtering strategy on the predictions (e.g., confidence thresholds), we do explore different strategies of \textit{a-priori} data selection, from two corpora: raw tweets (Section~\ref{sec:normSystem}), and the English Web Treebank~\cite{petrov2012overview}. 

For the English Web Treebank (EWT), we directly use raw text. Moreover, because the texts in the EWT are distributed by domains, i.e., \textit{answers}, \textit{emails}, \textit{weblogs}, \textit{newsgroups}, and \textit{reviews}, we preserve this information and keep the data separate according to their domain to see whether adding data from the different domains can provide a more useful signal.

For the raw tweets, we compare different strategies of sampling. In addition to selecting random tweets, we experimented with selections aimed at providing the tagger with specific information that we knew was missing or confusing in the original training data. One strategy thus was to include tweets that contained words occurring in the development data but not in the training data. 
Note that this would result in a very slow tagger in a real-world situation, since the tagger needs to be retrained for every new unknown word. 
Another strategy was based on a preliminary analysis of errors on the development data: from the confusion matrix we observed that a frequently confounded tag was proper noun. Considering named entities as adequate proxies for proper nouns in this context, we also experimented with adding tweets that contained named entities. The detection of named entities was performed using a Twitter-specific named entity recognizer~\cite{Ritter11}. For control and comparison, we also collect additional training data where only tweets that do \textit{not} contain named entities are selected. Hence, we end up with the following four sampling strategies:

\begin{itemize}
	\item random sampling
	\item tweets containing words which occur in the development data, but not in the training data
	\item tweets containing named entities 
	\item tweets not containing named entities
\end{itemize}

\paragraph{Results}

Adding more automatically-labeled data did not show any consistent improvement. This holds for both selection methods regarding named entities (presence/absence of NERs) and different domains of the Web treebank. Therefore we do not elaborate further  here.
We hypothesize that post-selection based on e.g., confidence sampling, is a more promising direction. We consider this future work.

\section{Normalizing and Not Normalizing}

\label{sec:analysis}

In the previous sections, we explored ways to improve the POS tagging of Tweets. 
The most promising directions were initializing the tagger with pre-trained embeddings and using normalization. Self-training was not effective. In this Section, we report on additional experiments on the development data aimed at obtaining insights on the potential of combining these two strategies.

\subsection{Consequences of Normalization}

\begin{table}[ht!]
\centering
\resizebox{\columnwidth}{!}{
	\begin{tabular}{l | r | r r r } \toprule
    				& \textsc{Bilty} & \textsc{+Norm} & \textsc{+Vecs}	& \textsc{+Comb} \\
                   
        \midrule
    	\textsc{canonical}	& 86.1	& 85.6	& 91.2 	& 90.1 \\
        \textsc{non-canon.} & 50.8	& 70.3	& 71.1 	& 78.5 \\
        \bottomrule
    \end{tabular}
    }
    \caption{Effect of different models on canonical/non-canonical words.}
    \label{tab:perfNormed}
\end{table}

\noindent Table~\ref{tab:perfNormed} shows the effect of the two approaches on the two subsets of tokens (canonical/non-canonical) on the \textsc{Dev} set. Word embeddings have a higher impact on standard, canonical tokens. It is interesting to note that word embeddings and normalization both have a similar yet complementary effect on the words to be normalized (non-canonical).
The improvements on non-canonical words seem to be complementary. The combined model additionally improves on words which need normalization, whereas it scores almost 1\% lower on canonical words.
This suggests that both strategies have potential to complement each other.

\subsection{Performance per POS}

\begin{figure}
	\centering
	\includegraphics[width=1.15 \columnwidth]{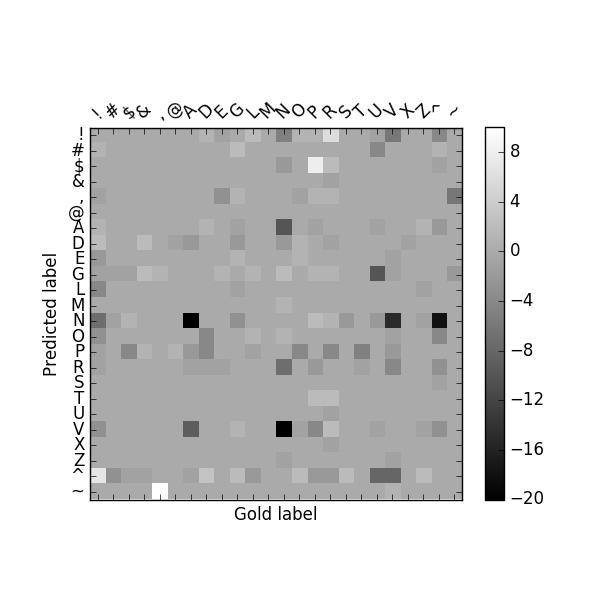}
	\caption{Differences in numbers of errors on development data between best normalization setting and best word embeddings. Dark means normalization makes more errors.}
    \label{fig:conf-best}
\end{figure}

We compare the type of errors made by the best normalization setting versus the best word embeddings setting in a confusion matrix which displays the difference in errors in Figure~\ref{fig:conf-best}. 
To recall: the best normalization setting was to use the raw training data, normalizing all words at test time; the best word embeddings model was a structured skip gram embeddings model with a window of 1. 

In the confusion graph it becomes clear that normalization results in over-predicting nouns (N), which often get confused with verbs (V), adjectives (A) and proper nouns (\^{}).
Normalization is better at recognizing prepositions (P), which it confuses less with numerals (\$) compared to the embedding model. This is due to normalizing `2' and `4'. Instead, the embedding model has better predictions for proper nouns, nouns and verbs, presumably due to the higher coverage.

\section{Evaluation}
\label{sec:eval}

In this section we report results on the test data, as introduced in Section~\ref{sec:data}. 

Our main aim is to compare different approaches for successfully applying a generic state-of-the-art POS tagger to Twitter data. Therefore we have to 
assess the contribution of the two methods we explore (normalization and using embeddings) and see how they fare, not only to each other but also in comparison to a state-of-the-art Twitter tagger. We use the ARK tagger~\cite{owoputi-EtAl:2013:NAACL-HLT} and retrain it on our dataset for direct comparison with our models.  The ARK system is a conditional random fields tagger, which exploits clusters, lexical features and gazetteers.

Table~\ref{tab:test} shows the performance of our best models and the ARK tagger on the test datasets.  

Embeddings work considerably better than normalization, which confirms what we found on the \textsc{Dev} data. The combined approach yields the highest accuracy over all evaluation sets, however, it significantly differs from embeddings only on \textsc{Test\_L}. This can be explained by our earlier observation (cf.\ Table~\ref{tab:perfNormed}), which shows that \textsc{Comb} yields the highest improvement on non-canonical tokens, but the same does not hold for canonical tokens. Notice that \textsc{Test\_L} does indeed contain the highest proportion of non-canonical tokens.

\begin{table*}
\centering
	\begin{tabular}{l | c c c } \toprule
 				& \textsc{Dev}  & \textsc{Test\_O} & \textsc{Test\_L} \\
                \midrule
         \% non-canonical tokens & 11.75\% & 10.95\% & 12.09\% \\
    	\midrule
		 \textsc{Bilty}	& 82.16 ($\pm$.33) & 83.81 ($\pm$.23) & 80.78 ($\pm$.32)\\
         \textsc{+Norm}	& 84.06 ($\pm$.32) & 84.73 ($\pm$.19) & 84.61 ($\pm$.21)\\
         \textsc{+Embeds}	& \textbf{88.51} ($\pm$.24) & \textbf{90.02} ($\pm$.35) & 88.53 ($\pm$.41)\\
         \textsc{+Comb}	& \textbf{88.89} ($\pm$.25)	& \textbf{90.25} ($\pm$.19) &  \textbf{89.63} ($\pm$.13)\\
    
        \midrule
         \textsc{Ark}		& 89.08			& 90.65 			& 89.67 \\
         
        \bottomrule
	\end{tabular}
    \caption{Results on the test data compared to ARK-tagger~\cite{owoputi-EtAl:2013:NAACL-HLT}. Bold: best result(s) for \textsc{Bilty} (bold: not significantly different from each other according to randomization test).}
    \label{tab:test}
\end{table*}

\begin{figure}[ht]
\centering
	\includegraphics[width=1.15\columnwidth]{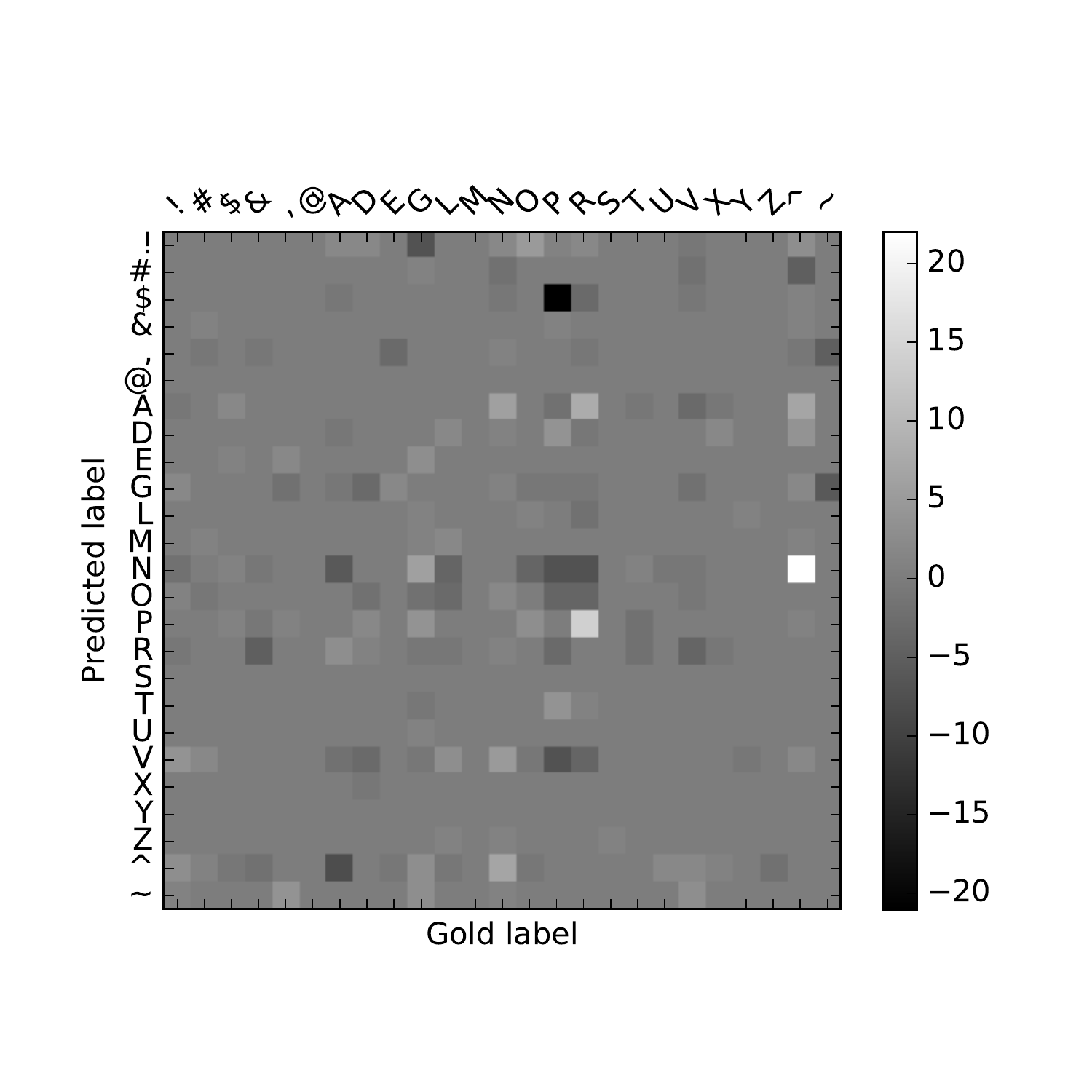}
	\caption{Comparison of errors per POS between our best model and the ARK tagger on \textsc{Test\_O}; darker means our system performs better.}
        \label{fig:combVSark} 
\end{figure}

Our best results on all datasets are comparable to the state-of-the-art results achieved by the ARK tagger. In Figure~\ref{fig:combVSark} we compare the errors made by our system (\textsc{Comb} in Table~\ref{tab:test}) and ARK on \textsc{Test\_O}, which is the test set on which both taggers obtain the highest performance.

The ARK tagger has difficulties with prepositions (P), which are mistagged as numerals (\$). These are almost all cases of `2' and `4', which represent Twitter slang for `to' and `for', respectively. Our system performs a lot better on these, due to the normalization model as already observed earlier. Still regarding prepositions, ARK is better at distinguishing them from adverbs (R), which is a common mistake for our system.
Our tagger makes more mistakes on confusing proper nouns (\^{}) with nouns (N) in comparison to ARK.

\section{Related Work}

Theoretically, this works fits well within the debate on normalization vs domain adaptation \cite{eisenstein:2013:NAACL-HLT}. For a practical comparison, the work most related to ours is that of~\citet{li2015joint}. 
They propose a joint model for normalization and POS
tagging. The candidate lists of six different normalization models, including spell checkers and machine translation systems, are combined with all their possible POS tags as found by the ARK Twitter POS tagger. Note that they use gold error detection, while we perform fully automatic normalization. These combined units of words and POS tags are then used to build joint Viterbi decoding~\cite{Viterbi1973Error}. The optimal path in this decoding does not only contain a sequence of normalized tokens, but also a sequence of POS tags. This joint model proves to be beneficial for both tasks.

Work on normalization for improving POS tagging has also been done on other languages. For example,  \citet{ljubevsic-erjavec-fivser:2017:BSNLP} show that performing normalization, in addition to using external resources, can remove half of the errors of a standard POS tagger for South Slavic languages. Quite surprisingly, instead, of all systems participating in shared tasks on POS tagging of Twitter data for both Italian \cite{bosco2016overview} and German \cite{beisswenger2016empirist}, none of the participating systems incorporated any normalization strategy before performing POS tagging.

Finally, normalization for POS tagging is certainly not limited to non-canonical data stemming from social media. Indeed, another stream of related work is focused on historical data, usually originating from the 15th till the 18th century. The motivation behind this is that in order to apply current language processing tools, the texts need to be normalized first, as spelling has changed through time. Experiments on POS tagging historical data that was previously normalized have been investigated for English \cite{yang-eisenstein:2016:N16-1}, German \cite{Bollmann13}, and Dutch~\citep{HUPKES16.196,tks2016chdh}. In this latter work, different methods of `translating' historical Dutch texts to modern Dutch are explored, and 
a vocabulary lookup-based approach appears to work best.\footnote{Interestingly, this work also resulted in a shared task on normalization of historical Dutch, in which the secondary evaluation metric was POS tagging accuracy: \url{https://ifarm.nl/clin2017st/}.} In this paper we focused on normalization and POS tagging for Twitter data only.

\section{Conclusions}

We investigated the impact of normalization on POS tagging for the Twitter domain, presenting the first results on automatic normalization and comparing normalization to alternative strategies. We compared a generic tagger to a tagger specifically designed for Twitter data.

Regarding Q1, we can conclude that normalization does help. However, using large amounts of unlabeled data for embedding initialization yields an improvement that is twice as large as the one obtained using normalization (Q2).

Combining both methods (Q3) does indeed yield the highest scores on all datasets.
This suggests that the two approaches are complementary, also because in isolation their most frequent errors differ.
However, the contribution of normalization on top of embeddings alone is relatively small and only significant on one test set, which was specifically developed for normalization and contains the largest proportion of non-canonical tokens. 

Overall, our best model is comparable to the ARK tagger. As a general direction, our results suggest that exploiting large amounts of unlabeled data of the target domain is preferable. However, if the data is expected to include a large proportion of non-canonical tokens, it is definitely worth applying normalization in combination with embeddings.  

Our investigation was limited by the amount of available training data. Adding data via self-training did not help.  We observed mixed results for different types of a-priori filtering, but none of them yielded a steady improvement. A more promising direction might be post-selection, based on confidence scores or agreement among different taggers. Obviously another way to go is to add manually labeled data, some of which is available for more canonical domains. This would require a mapping of tagsets, and might be another good testbed to assess the contribution of normalization, which we leave for future work.

All code and distributable data used in this paper is available at \url{https://github.com/bplank/wnut-2017-pos-norm}.

\section*{Acknowledgments}
We want to thank H\'{e}ctor Mart\'{i}nez Alonso and Gertjan van Noord for valuable comments on earlier drafts of this paper. We are also grateful to the anonymous reviewers. This research has been supported by the Nuance Foundation, NVIDIA research and the University of Groningen High Performance Computing center.

\bibliography{emnlp2017}
\bibliographystyle{emnlp_natbib}

\end{document}